\documentclass[10pt,twocolumn,letterpaper]{article}

\usepackage[pagenumbers]{cvpr} 

\definecolor{cvprblue}{rgb}{0.21,0.49,0.74}
\usepackage[pagebackref,breaklinks,colorlinks,allcolors=cvprblue]{hyperref}

\usepackage{amsmath}
\usepackage{amssymb}
\usepackage{mathtools}
\usepackage{amsthm}
\usepackage{multirow}
\usepackage{colortbl}
\usepackage{adjustbox}

\usepackage[table]{xcolor}
\usepackage{xcolor}
\usepackage{booktabs}
\usepackage{tabularx}
\usepackage{makecell}
\newcolumntype{C}{>{\centering\arraybackslash}X}

\usepackage{siunitx}

\def\confName{CVPR}
\def\confYear{2027}

\title{Contrast Where It Matters: Visual Evidence-Guided Decoding for Mitigating Hallucinations in LVLMs}

\makeatletter
\newcommand{\myfnsymbol}[1]{%
  \expandafter\@myfnsymbol\csname c@#1\endcsname
}
\newcommand{\@myfnsymbol}[1]{%
  \ifcase #1
  \or \TextOrMath{\textasteriskcentered}{*}
  \or \TextOrMath{\textdagger}{\dagger}
  \fi
}
\newcommand{\equalcontributor}{\@myfnsymbol{1}}
\newcommand{\corresponding}{\@myfnsymbol{2}}
\makeatother

\author{
Yiheng Li$^{1,2}$,
Wenhao Wang$^{3,\corresponding}$,
Xingyi Li$^{4}$,
Mingda Jia$^{1,2}$,
Zichang Tan$^{6}$,\\
Yinhang Tang$^{7}$,
Zhen Lei$^{1,2,5,8}$,
Yang Yang$^{1,2,\corresponding}$\\
$^1$ School of Artificial Intelligence, University of Chinese Academy of Sciences\\
$^2$ MAIS, Institute of Automation, Chinese Academy of Sciences\\
$^3$ Vast Intelligence Lab,
$^4$ Institute of Information Engineering, Chinese Academy of Sciences\\
$^5$ CAIR, HKISI, Chinese Academy of Sciences, $^6$ Sangfor Technologies Inc.\\
$^7$ AInnovation,
$^8$ SCSE, FIE, M.U.S.T\\
{\tt\small \{liyiheng2024, yangyang2013, zhen.lei\}@ia.ac.cn, wangwenhao@vastilab.com} 
}

\begin{document}
\maketitle
\renewcommand{\thefootnote}{\myfnsymbol{footnote}}
\footnotetext[2]{Corresponding authors}%
\begin{abstract}
Despite strong multimodal capabilities, large vision-language models (LVLMs) often hallucinate when linguistic priors outweigh visual evidence or when predictions are grounded in irrelevant or misleading image regions.
Training-free contrastive decoding mitigates hallucinations by contrasting the original prediction with an auxiliary one that exposes linguistic priors.
Yet existing methods focus on altering the auxiliary input or computation and do not ensure that either branch uses query-relevant visual evidence during decoding.
We propose Visual Evidence-Guided Decoding (VEGD), a training-free framework that identifies query-relevant visual evidence and guides its use throughout contrastive decoding.
VEGD locates this evidence by aggregating localization heads with strong visual responses and coherent spatial structure, then selectively weakens it in the auxiliary image.
During generation, VEGD directs visual-carrier heads in both branches toward the same evidence location, strengthening access to intact evidence in the main branch while anchoring the auxiliary branch to its weakened counterpart.
An adaptive distribution fusion strategy then combines or contrasts their predictions according to distribution divergence.
Experiments across diverse LVLMs and benchmarks demonstrate that VEGD consistently reduces hallucinations.
\end{abstract}    
\section{Introduction}
\label{sec:intro}

\begin{figure}[t]
    \centering
    \includegraphics[width=0.46\textwidth]{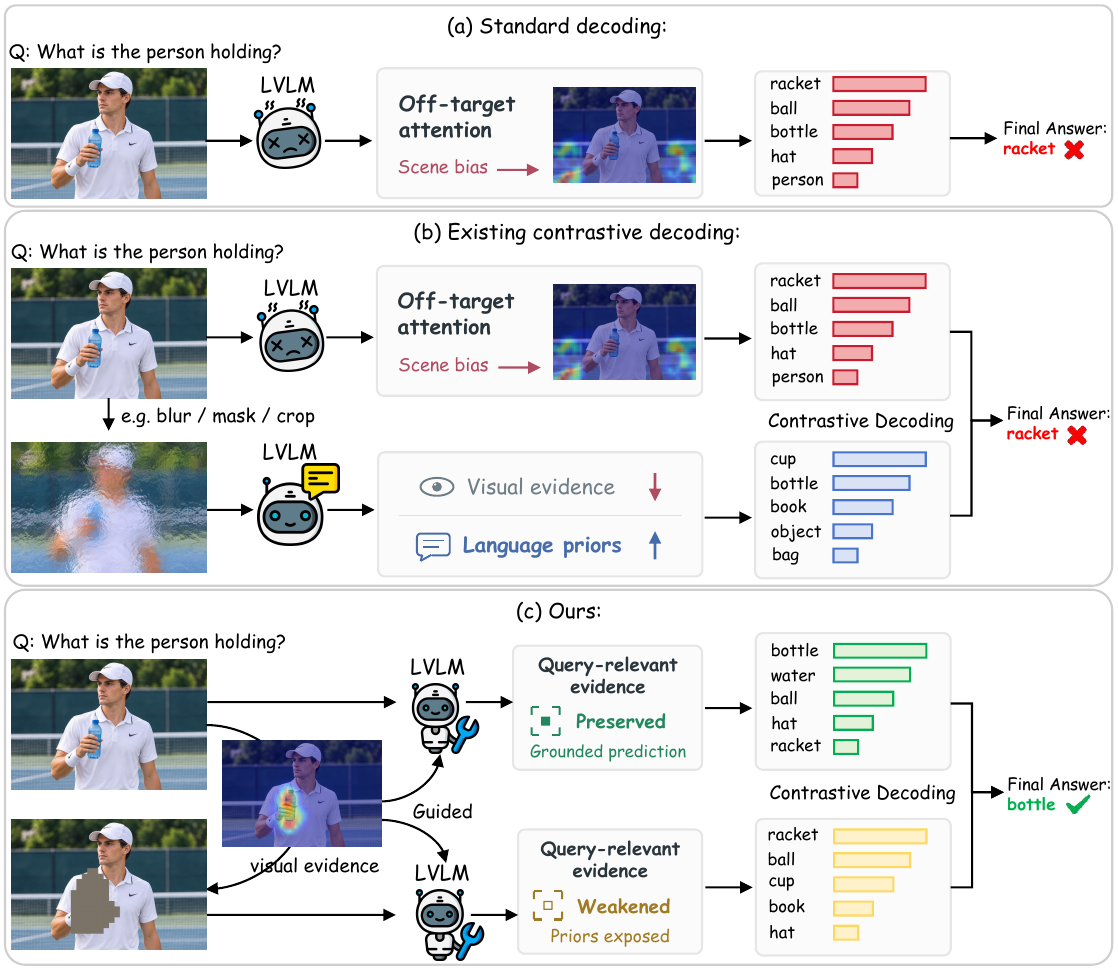}
    \caption{
    \textbf{Comparison of visual evidence utilization between our method and existing decoding methods.}
    (a) Standard decoding is distracted by irrelevant visual content.
    (b) Existing contrastive decoding exposes language priors through the auxiliary branch, but the main branch may still rely on irrelevant or misleading visual evidence.
    (c) Our method identifies query-relevant visual evidence, uses it to construct the auxiliary input, and guides both branches toward it throughout decoding.
    }
    \label{pic0}
\end{figure}

Large vision-language models (LVLMs) have made remarkable progress in multimodal understanding and generation, yet they remain prone to visual hallucinations~\cite{liu2023visual,dai2023instructblip,li2023evaluating,leng2024mitigating}, especially when learned linguistic priors outweigh visual evidence~\cite{favero2024multimodal,leng2024mitigating,liu2024paying}.
As shown in Fig.~\ref{pic0}(a), these errors also occur when the model relies on irrelevant or misleading image regions and grounds its prediction in the wrong visual evidence~\cite{woo2024attentional,an2025agla,li2026cmac}.
Reliable predictions must therefore remain faithful to the image rather than simply follow plausible linguistic or contextual patterns.

To mitigate hallucinations without additional training, recent work has explored several forms of inference-time intervention.
Some methods calibrate output distributions~\cite{favero2024multimodal,leng2024mitigating}, while others steer internal representations~\cite{liu2025vti,li2025hidden} or modify model computation~\cite{wan2025only,li2026cmac,lei2026see}.
Among these approaches, contrastive decoding calibrates token probabilities by comparing the original output distribution with an auxiliary one.
Tokens that remain likely in the auxiliary branch are viewed as more dependent on linguistic priors, whereas those favored by the original input are promoted~\cite{leng2024mitigating,lee2026delve}.
Early methods constructed the auxiliary distribution through global image corruption or transformation~\cite{leng2024mitigating,woo2024ritual,lee2026delve}, while later work modified selected regions or visual tokens~\cite{an2025agla,woo2024attentional,huang2026chasd,liang2026envision}.
Related approaches instead create the altered prediction inside the model~\cite{li2026cmac,wan2025only}. Despite their different implementations, these methods calibrate predictions through the output gap between the original and altered branches.

Contrastive decoding can suppress prior-driven predictions, but it does not control which visual evidence the original and auxiliary branches use during generation, as shown in Fig.~\ref{pic0}(b).
Global methods perturb the whole image, so changes unrelated to the query can affect the output difference~\cite{leng2024mitigating,woo2024ritual,lee2026delve}.
Local methods restrict the perturbation to selected regions or visual tokens~\cite{an2025agla,woo2024attentional,huang2026chasd,liang2026envision}, but their localization cues typically determine only where the auxiliary input is altered~\cite{an2025agla,huang2026chasd,liang2026envision}.
Such cues neither direct the original branch toward query-relevant evidence nor anchor the auxiliary branch to the corresponding weakened region during decoding.
Methods that form the altered prediction inside the model avoid explicit input perturbation, but their token- or attention-level interventions likewise do not explicitly coordinate the visual evidence used by the two branches~\cite{huo2025self,li2026cmac,wan2025only}.

Closing this gap requires locating query-relevant visual evidence and keeping both branches focused on it throughout contrastive decoding.
Identifying this evidence should not rely on averaging all attention heads, since visual grounding is concentrated in a few localization heads and indiscriminate aggregation mixes their spatial signal with noisy or non-grounding responses~\cite{kang2025your,xie2026headaware}.
As illustrated in Fig.~\ref{pic0}(c), guiding the main branch toward intact evidence helps the model favor what is actually depicted over predictions suggested by language priors or surrounding content.
In the auxiliary branch, the same evidence is weakened while attention remains anchored to its location, exposing prior-driven predictions without allowing the branch to seek support elsewhere.
Contrasting the two distributions can therefore distinguish visually supported predictions from prior-driven alternatives, making query-relevant evidence a persistent decoding guide rather than a one-off cue for constructing the auxiliary input.

Based on this principle, we propose Visual Evidence-Guided Decoding (VEGD), a training-free framework that localizes query-relevant visual evidence and guides its use across both branches.
VEGD selects localization heads by visual response and spatial coherence, then aggregates their maps to locate the evidence region.
It selectively weakens this region to construct the auxiliary image while preserving the rest of the scene.
During decoding, the same location guides visual-carrier heads in both branches, reinforcing intact evidence in the main branch and anchoring the auxiliary branch to its weakened counterpart.
Finally, VEGD combines the two output distributions according to their discrepancy, preserving complementary predictions when they agree and emphasizing their contrast when they diverge.
On CHAIR~\cite{rohrbach2018object}, POPE~\cite{li2023evaluating}, and MME-Hallucination~\cite{fu2025mme} across multiple LVLMs, VEGD consistently reduces hallucinations. Our contributions are:
\begin{itemize}
    \item We identify visual-evidence use during decoding as a key factor in contrastive hallucination mitigation: refining the auxiliary branch alone is insufficient when either branch can still be distracted by irrelevant or misleading regions.
    \item We propose training-free VEGD, which localizes query-relevant visual evidence with selected heads, weakens it in the auxiliary image, and anchors both branches to it during decoding.
    \item Experiments on three benchmarks across multiple LVLMs demonstrate consistent hallucination reduction across established and recent model families.
\end{itemize}

\section{Related Work}
\label{sec:related}

\subsection{Large Vision-Language Models}
Large vision-language models (LVLMs) have advanced rapidly in open-ended visual understanding and reasoning. Early work focused on connecting pretrained visual and language components. Flamingo~\cite{alayrac2022flamingo} introduced gated cross-attention, while BLIP-2~\cite{li2023blip2} and InstructBLIP~\cite{dai2023instructblip} used a lightweight Querying Transformer. MiniGPT-4~\cite{zhu2024minigpt4} used lightweight visual alignment, while LLaVA~\cite{liu2023visual}, ShareGPT4V~\cite{chen2024sharegpt4v}, and Shikra~\cite{chen2023shikra} advanced visual instruction tuning for dialogue and grounding. More recent models~\cite{li2024llavaonevision,bai2025qwen25vl,zhu2025internvl3} scale multimodal pretraining to higher-resolution images, multiple images, videos, and broader reasoning tasks. Despite this progress, hallucinations remain common~\cite{li2023evaluating,leng2024mitigating}, motivating mitigation methods that transfer across architectures.

\begin{figure*}[t]
    \centering
    \includegraphics[width=0.98\textwidth]{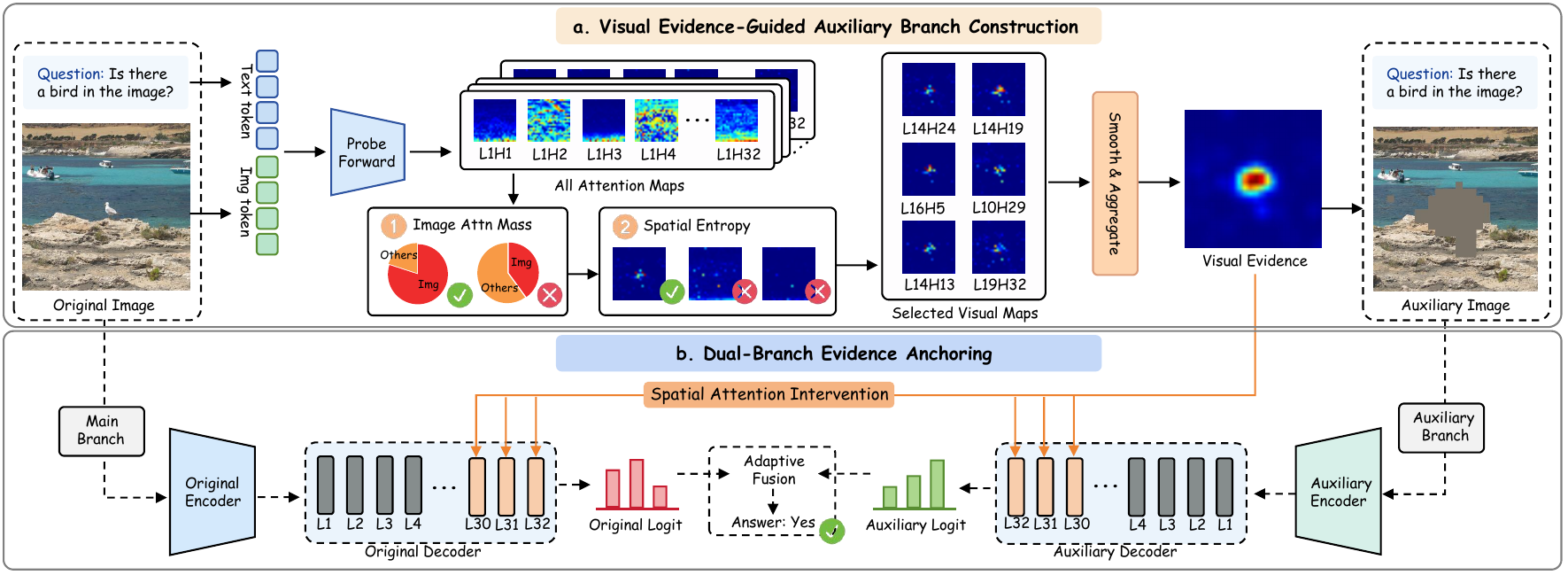}
    \caption{
    \textbf{Overview of Visual Evidence-Guided Decoding.}
    (a) Visual Evidence-Guided Auxiliary Branch Construction identifies query-relevant visual evidence with selected localization heads and uses it to construct the auxiliary image.
    (b) Dual-Branch Evidence Anchoring guides visual-carrier heads in both branches toward the same evidence during decoding, followed by adaptive fusion of their predictions.
    }
    \label{fig:overview}
\end{figure*}

\subsection{Contrastive Decoding for LVLM Hallucination}
Visual hallucination occurs when an LVLM generates content unsupported by the image, including nonexistent objects and incorrect attributes~\cite{rohrbach2018object,li2023evaluating,wang2024instruction}. Contrastive decoding originally selects tokens favored by a strong expert over a weaker amateur model~\cite{li2023contrastive}. LVLM methods adapt this idea by contrasting the standard prediction with an auxiliary distribution in which visual grounding is weakened or altered. Early variants modify the conditioning signal to create this counterpart. VCD~\cite{leng2024mitigating} adds image noise, RITUAL~\cite{woo2024ritual} applies image transformations, and Instruction Contrastive Decoding~\cite{wang2024instruction} perturbs the instruction. M3ID~\cite{favero2024multimodal} and PAI~\cite{liu2024paying} use language-only predictions, while DeGF~\cite{zhang2025selfcorrecting} uses a generated image for feedback. Later methods make the auxiliary prediction more targeted by modifying selected regions or visual tokens. AGLA~\cite{an2025agla} and EnAR~\cite{liang2026envision} use local masking, AvisC~\cite{woo2024attentional} biases predictions toward blind tokens, and CHASD~\cite{huang2026chasd} perturbs salient visual tokens. A third group constructs the contrast inside the network rather than corrupting the input. SID~\cite{huo2025self} and CMAC~\cite{li2026cmac} intervene in visual tokens or cross-modal values, while ONLY~\cite{wan2025only} and ACG~\cite{jo2026attention} operate on internal layers or attention paths. The spatial or internal intervention, however, is typically used to create that prediction, while how both branches retrieve visual evidence during subsequent decoding remains less explored.

\subsection{Visual Evidence Localization in LVLMs}
Reliable use of visual evidence begins with locating regions relevant to the query. V*~\cite{wu2024vstar} and ViCrop~\cite{zhang2025mllms} expose fine-grained evidence through visual search and model-guided cropping, while F-LMM~\cite{wu2025flmm} shows that word--pixel correspondences emerge in LVLM attention. This signal is highly head-dependent. SparseMM~\cite{wang2025sparsemm} finds that few heads contribute strongly to visual understanding, localization-head analysis~\cite{kang2025your} identifies a small set that reliably recovers referred objects, and attention-lens analysis~\cite{jiang2025devils} links hallucinations to heads attending to inconsistent objects. Head-Aware Visual Cropping~\cite{xie2026headaware} likewise selects grounding heads before constructing its guidance map. These findings make selected-head attention a training-free source for locating query-relevant visual evidence. VEGD builds on this insight to construct the auxiliary view and guide both branches with the same localized evidence throughout decoding.

\section{Method}
\label{sec:method}

\begin{figure*}[t]
    \centering
    \includegraphics[width=0.98\textwidth]{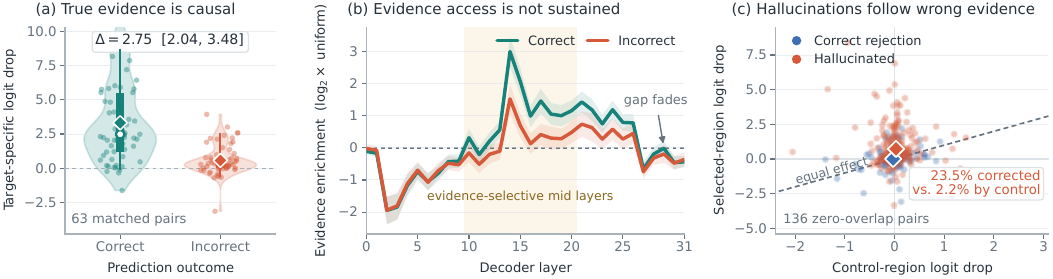}
    \caption{\textbf{Visual-evidence analysis.} (a) Correct predictions depend more strongly on query-relevant evidence. (b) Direct access to this evidence strengthens in middle layers but weakens later. (c) Hallucinations can instead follow a causally misleading visual region. Bands in (b) denote 95\% bootstrap confidence intervals.}
    \label{fig:motivation-analysis}
\end{figure*}

\subsection{Preliminaries and Motivating Analysis}
\label{sec:preliminaries}

\noindent\textbf{Preliminaries.}
We consider an autoregressive LVLM that represents an image $I$ as $N_v$ visual tokens $\mathbf{X}^{v}\in\mathbb{R}^{N_v\times d}$ and embeds a query $Q$ as $N_q$ text tokens $\mathbf{X}^{q}\in\mathbb{R}^{N_q\times d}$, where $d$ is the shared embedding dimension. At generation step $t$, the decoder processes:
\begin{equation}
    \mathbf{X}_t=\operatorname{Concat}
    \bigl(\mathbf{X}^{v},\mathbf{X}^{q},\mathbf{X}^{<t}\bigr),
\end{equation}
where $\mathbf{X}^{<t}$ contains the embeddings of previously generated tokens. Let $\mathcal{I}_v$ denote the visual-token positions in $\mathbf{X}_t$.

Let $L$ and $H$ denote the numbers of decoder layers and attention heads per layer, respectively. For layer $\ell\in\{1,\ldots,L\}$ and head $h\in\{1,\ldots,H\}$, self-attention is computed as:
\begin{equation}
    \small
    \mathbf{A}^{\ell,h}
    =\operatorname{softmax}\!\left(
    \frac{\mathbf{Q}^{\ell,h}(\mathbf{K}^{\ell,h})^{\top}}{\sqrt{d_h}}\right),
    \mathbf{O}^{\ell,h}=\mathbf{A}^{\ell,h}\mathbf{V}^{\ell,h},
    \label{eq:self_attention}
\end{equation}
Here, $\mathbf{Q}^{\ell,h}$, $\mathbf{K}^{\ell,h}$, and $\mathbf{V}^{\ell,h}$ are the query, key, and value matrices, $d_h$ is the head dimension, and $\mathbf{A}^{\ell,h}$ and $\mathbf{O}^{\ell,h}$ are the attention matrix and head output. The visual attention from query position $r$ is:
\begin{equation}
    \mathbf{a}^{\ell,h}_{r}
    =\mathbf{A}^{\ell,h}_{r,\mathcal{I}_v}\in\mathbb{R}^{N_v},
    \qquad
    m^{\ell,h}_{r}=\sum_{j=1}^{N_v}\mathbf{a}^{\ell,h}_{r,j}.
    \label{eq:visual_attention}
\end{equation}
Here, $m^{\ell,h}_{r}$ measures how strongly the head reads visual information. Let $H_v\times W_v$ denote the spatial layout of the visual tokens, where $N_v=H_vW_v$. Reshaping $\mathbf{a}^{\ell,h}_{r}$ accordingly gives the head-wise spatial map $\mathbf{S}^{\ell,h}_{r}\in\mathbb{R}^{H_v\times W_v}$ used below.
The vocabulary projection $\mathbf{W}_{o}$ maps the final-layer state $\mathbf{h}^{L}_{t}$ to logits $\mathbf{z}_t=\mathbf{W}_{o}\mathbf{h}^{L}_{t}$ and distribution $\mathbf{p}_t=\operatorname{softmax}(\mathbf{z}_t)$ over $\mathcal{V}$. VEGD uses these head-wise visual maps and output distributions.

\noindent\textbf{Motivating analysis.}
Using the logit difference between Yes and No at the first answer position, we analyze LLaVA-1.5-7B~\cite{liu2023visual} on adversarial POPE-COCO~\cite{li2023evaluating}.
The diagnostic set contains all 1,500 negative queries and 1,263 positive queries with target boxes visible after preprocessing.

\noindent\textbf{Correct predictions rely on relevant evidence.}
Of 1,263 positive queries, the model predicts 1,200 correctly and 63 incorrectly.
We match each error with a correct prediction for the same queried category and the closest target-object area.
For every example, we mask either the annotated target or an equal-area control region and record the change in the Yes--No logit margin.
We define the target-specific effect as the margin drop under the target mask minus that under the control mask.
In Fig.~\ref{fig:motivation-analysis} (a), the mean effect is 3.32 for correct predictions but only 0.57 for errors, giving a paired gap of 2.75 (95\% CI: [2.04, 3.48]).
Successful predictions thus draw much stronger causal support from the query-relevant region.

\noindent\textbf{Direct evidence access is strongest in middle layers.}
Using the same pairs, we track how access to the target evidence changes across the decoder.
At each layer, area-normalized evidence enrichment compares the target region's share of visual attention with its share of image patches.
As Fig.~\ref{fig:motivation-analysis} (b) shows, correct and incorrect cases behave similarly in early layers but separate in the middle layers.
The paired gap in $\log_2$ evidence enrichment averages 0.79 over layers 10--20 (95\% CI: [0.43, 1.17]) but falls to 0.32 over layers 21--31 (95\% CI: [0.03, 0.61]).
The model can therefore access relevant evidence in the middle layers, yet this advantage largely fades toward the output.

\noindent\textbf{Hallucinations can follow the wrong evidence.}
Finally, we test whether a different object in the image can drive a hallucinated answer.
Among 1,500 negative queries, the model incorrectly predicts Yes in 453 cases.
Before masking anything, we select the present-object region with the strongest middle-layer enrichment and match each hallucination with a correct rejection by queried category, candidate area, and the number of present object categories.
Fig.~\ref{fig:motivation-analysis} (c) reports a strict subset of 136 pairs whose equal-area control masks overlap no annotated object.
The candidate-specific effect averages 0.75 for hallucinations and 0.02 for correct rejections.
Masking the candidate region corrects 23.5\% of hallucinations, whereas masking the control corrects only 2.2\%.
Some hallucinations are therefore not image-independent guesses: the model reads the image but follows the wrong evidence.

These findings lead to two design choices. First, because correct predictions depend on query-relevant regions, we mask these regions in the auxiliary image. This weakens the relevant evidence while leaving the rest of the scene largely unchanged, giving a cleaner comparison with the original image. Second, the model attends to the right evidence most strongly in the middle layers, but this focus fades toward the output and may shift to a misleading region. We therefore reuse the localized evidence map to guide attention in the final layers of both branches. As shown in Fig.~\ref{fig:overview}, the same map is used first to construct the auxiliary image and then to guide later decoding.

\subsection{Visual Evidence-Guided Auxiliary Branch Construction}
\label{sec:auxiliary-construction}

Fig.~\ref{fig:overview} (a) shows how VEGD constructs the auxiliary branch. Given an image $I$ and a query $Q$, one probe forward pass collects the head-wise spatial maps $\mathbf{S}_{r_q}^{\ell,h}$ at the last query position $r_q$. The module then selects localization heads, combines their maps into visual evidence $\mathbf{E}$, and uses $\mathbf{E}$ to produce the auxiliary image $I^a$. This process is training-free and runs once before generation.

\noindent\textbf{Localization-head selection.}
Not every head-wise map from the probe is a reliable localization cue. Attention maps can expose word--pixel correspondences~\cite{wu2025flmm}, but reliable grounding is concentrated in only a few heads~\cite{kang2025your,xie2026headaware}. Some heads place little mass on image tokens, while others scatter their response across unrelated regions. Guided by Fig.~\ref{fig:motivation-analysis} (b), we first limit candidates to the middle-layer set $\mathcal{L}_{\mathrm{loc}}$, where the gap in relevant-evidence access between correct and incorrect predictions is largest. We then apply two filters with distinct roles. For image attention mass, we rank heads by $m_{r_q}^{\ell,h}$ and retain those at or above the elbow of the resulting curve~\cite{kang2025your}. For each remaining head, we isolate the elevated response $\mathbf{R}^{\ell,h}=[\mathbf{S}_{r_q}^{\ell,h}-2\bar{S}_{r_q}^{\ell,h}]_{+}$ and group its nonzero locations into connected components $\mathcal{C}^{\ell,h}$. We measure their spatial fragmentation $\eta^{\ell,h}$ by:
\begin{equation}
    \eta^{\ell,h}
    =-\sum_{c\in\mathcal{C}^{\ell,h}}p_c^{\ell,h}\log p_c^{\ell,h},
    \quad
    p_c^{\ell,h}=\frac{\sum_{u\in c}\mathbf{R}_u^{\ell,h}}{\sum_u\mathbf{R}_u^{\ell,h}}.
    \label{eq:spatial-entropy}
\end{equation}
Here, $p_c^{\ell,h}$ is the fraction of retained response in component $c$, $\bar{S}_{r_q}^{\ell,h}$ is the spatial mean, and $[\cdot]_+$ denotes the positive part. A lower $\eta^{\ell,h}$ means that the elevated attention is concentrated in fewer components. We retain the $K$ lowest-entropy heads as $\mathcal{H}_{\mathrm{loc}}$: the mass filter ensures a visual response, and the entropy filter gives that response a compact spatial support.

\noindent\textbf{Visual-evidence aggregation.}
The selected maps are smoothed with a Gaussian kernel $\mathcal{G}_{\sigma}$ to reduce isolated patch responses and are then averaged:
\begin{equation}
    \mathbf{E}
    =\frac{1}{|\mathcal{H}_{\mathrm{loc}}|}
    \sum_{(\ell,h)\in\mathcal{H}_{\mathrm{loc}}}
    \mathcal{G}_{\sigma}*\mathbf{S}_{r_q}^{\ell,h}.
    \label{eq:visual-evidence-map}
\end{equation}
Selection takes place before aggregation so that heads with weak or fragmented spatial responses do not blur the resulting map. The resulting $\mathbf{E}\in\mathbb{R}^{H_v\times W_v}$ assigns a continuous score to each visual location. This map is used to construct the auxiliary image and later guide both branches.

\noindent\textbf{Auxiliary-image construction.}
The evidence map $\mathbf{E}$ determines which part of the image should differ between the two branches. We threshold it at its mean score to obtain a binary patch mask $\mathbf{M}$ that marks locations with above-average evidence. We then resize $\mathbf{M}$ to the image resolution as $\widehat{\mathbf{M}}$ and construct the auxiliary image as:
\begin{equation}
    I^{a}=(1-\widehat{\mathbf{M}})\odot I
    +\widehat{\mathbf{M}}\odot\boldsymbol{\mu}.
    \label{eq:auxiliary-image}
\end{equation}
Here, $\boldsymbol{\mu}$ is the image-normalization mean. Replacing the selected region weakens the localized visual evidence while preserving the rest of the scene. The two images therefore differ mainly in the evidence being tested. VEGD retains $\mathbf{E}$ after constructing $I^a$ to guide both branches to the same location in the next stage.

\subsection{Dual-Branch Evidence Anchoring}
\label{sec:evidence-anchoring}

Fig.~\ref{fig:overview} (b) illustrates dual-branch evidence anchoring. The original image $I$ and auxiliary image $I^a$ are decoded in parallel using the same previously generated tokens. Let $\mathcal{L}_{\mathrm{anc}}$ denote the final decoder layers where anchoring is applied. At each generation step, the same evidence map $\mathbf{E}$ guides both branches in these layers, maintaining access to relevant evidence toward the output.

\noindent\textbf{Visual-carrier head selection.}
The heads used here serve a different role from the localization heads in Sec.~\ref{sec:auxiliary-construction}. Localization heads must provide a coherent spatial map, whereas visual-carrier heads only need to carry substantial visual information in the layer being intervened. We therefore score each head by the image attention mass $m_{r_q}^{\ell,h}$ from the probe pass. For each layer $\ell\in\mathcal{L}_{\mathrm{anc}}$, we rank these scores and retain heads at or above the elbow of the resulting curve~\cite{kang2025your}. Among them, we select the strongest $K_a$ heads, with $K_a$ setting the number selected per layer. We repeat this procedure independently across layers because different heads may carry visual information at different depths. The same selected heads are used in both branches.

\noindent\textbf{Spatial Attention Intervention.}
To carry the localized evidence into decoding, we align $\mathbf{E}$ with the visual-token sequence. We use the lower and upper deciles of the aligned scores, $q^{-}$ and $q^{+}$, as normalization bounds and compute the guidance scores $\mathbf{e}\in[0,1]^{N_v}$ as:
\begin{equation}
    e_j=\left(\Pi_{[0,1]}\!\left(
    \frac{\mathbf{E}_j-q^{-}}{\max(q^{+}-q^{-},\varepsilon)}
    \right)\right)^{\rho},
    \label{eq:normalized-evidence}
\end{equation}
where $\Pi_{[0,1]}$ projects a scalar onto $[0,1]$, $\rho$ controls how sharply the intervention concentrates on high-evidence tokens, and $\varepsilon=10^{-6}$ prevents division by zero. At each anchored layer, $e_j$ modulates the selected visual-carrier heads. For any selected head, let $g_{t,j}$ denote the corresponding pre-softmax attention logit at step $t$. Extending the logit amplification used by PAI~\cite{liu2024paying} with spatially varying guidance, we update it as:
\begin{equation}
    \widetilde{g}_{t,j}
    =g_{t,j}
    +\alpha\left|g_{t,j}\right|e_j,
    \qquad j\in\mathcal{I}_v,
    \label{eq:evidence-attention-bias}
\end{equation}
where $\alpha$ controls the intervention strength and can be set separately for the two branches. All other attention logits remain unchanged. In the main branch, the update reinforces intact query-relevant evidence. In the auxiliary branch, it keeps attention on the same location after its content has been weakened. This shared spatial guidance prevents either branch from drifting toward unrelated regions.

\definecolor{POPEGain}{rgb}{0.25,0.49,0.41}
\definecolor{POPELoss}{rgb}{0.71,0.37,0.37}
\newcommand{\popegain}[1]{\textcolor{POPEGain}{\tiny\,(+#1)}}
\newcommand{\popeloss}[1]{\textcolor{POPELoss}{\tiny\,(-#1)}}

\begin{table*}[t]
\centering

\caption{POPE~\citep{li2023evaluating} results on MS-COCO, A-OKVQA, and GQA, averaged over the random, popular, and adversarial splits. Acc. and Prec. denote accuracy and precision. Colored values show absolute changes from Regular decoding.}
\label{tab:POPE}

\begingroup
\scriptsize
\setlength{\tabcolsep}{1.8pt}
\renewcommand{\arraystretch}{0.94}

\begin{tabularx}{\linewidth}{
    >{\centering\arraybackslash}p{0.62cm}
    >{\raggedright\arraybackslash}p{1.05cm}
    @{\hspace{2.2pt}}
    *{3}{>{\centering\arraybackslash}X}
    @{\hspace{1.8pt}}
    !{\color{black!25}\vrule width 0.45pt}
    @{\hspace{1.8pt}}
    *{3}{>{\centering\arraybackslash}X}
    @{\hspace{1.8pt}}
    !{\color{black!25}\vrule width 0.45pt}
    @{\hspace{1.8pt}}
    *{3}{>{\centering\arraybackslash}X}
}

\toprule
&
\multirow{2}{*}{\textbf{Method}}
&
\multicolumn{3}{c}{\textbf{LLaVA-1.5-7B}}
&
\multicolumn{3}{c}{\textbf{InstructBLIP-7B}}
&
\multicolumn{3}{c}{\textbf{ShareGPT4V-7B}}
\tabularnewline

\cmidrule(lr){3-5}
\cmidrule(lr){6-8}
\cmidrule(lr){9-11}

&
&
\mbox{Acc. $\uparrow$}
&
\mbox{Prec. $\uparrow$}
&
\mbox{F1 $\uparrow$}
&
\mbox{Acc. $\uparrow$}
&
\mbox{Prec. $\uparrow$}
&
\mbox{F1 $\uparrow$}
&
\mbox{Acc. $\uparrow$}
&
\mbox{Prec. $\uparrow$}
&
\mbox{F1 $\uparrow$}
\tabularnewline

\midrule

\multirow{5}{*}{\rotatebox[origin=c]{90}{\textbf{MS-COCO}}}
& Regular
& $80.58$
& $77.40$
& $81.84$
& $77.85$
& $74.85$
& $79.39$
& $81.58$
& $79.50$
& $82.35$
\tabularnewline

& VCD~\cite{leng2024mitigating}
& $80.22$ \popeloss{0.36}
& $76.67$ \popeloss{0.73}
& $81.69$ \popeloss{0.15}
& $79.27$ \popegain{1.42}
& $77.19$ \popegain{2.34}
& $80.29$ \popegain{0.90}
& $81.91$ \popegain{0.33}
& $80.24$ \popegain{0.74}
& $82.57$ \popegain{0.22}
\tabularnewline

& AvisC~\cite{woo2024attentional}
& $83.36$ \popegain{2.78}
& $81.00$ \popegain{3.60}
& $84.23$ \popegain{2.39}
& $\underline{84.33}$ \popegain{6.48}
& $\underline{85.19}$ \popegain{10.34}
& $\underline{84.32}$ \popegain{4.93}
& $\underline{85.72}$ \popegain{4.14}
& $\underline{86.28}$ \popegain{6.78}
& $\underline{85.73}$ \popegain{3.38}
\tabularnewline

& ONLY~\cite{wan2025only}
& $\underline{84.88}$ \popegain{4.30}
& $\underline{83.09}$ \popegain{5.69}
& $\underline{85.51}$ \popegain{3.67}
& $82.59$ \popegain{4.74}
& $79.21$ \popegain{4.36}
& $83.85$ \popegain{4.46}
& $83.77$ \popegain{2.19}
& $80.07$ \popegain{0.57}
& $84.93$ \popegain{2.58}
\tabularnewline

& \cellcolor{cvprblue!6}Ours
& \cellcolor{cvprblue!6}$\mathbf{85.92}$ \popegain{5.34}
& \cellcolor{cvprblue!6}$\mathbf{86.25}$ \popegain{8.85}
& \cellcolor{cvprblue!6}$\mathbf{86.03}$ \popegain{4.19}
& \cellcolor{cvprblue!6}$\mathbf{85.04}$ \popegain{7.19}
& \cellcolor{cvprblue!6}$\mathbf{87.62}$ \popegain{12.77}
& \cellcolor{cvprblue!6}$\mathbf{84.64}$ \popegain{5.25}
& \cellcolor{cvprblue!6}$\mathbf{86.02}$ \popegain{4.44}
& \cellcolor{cvprblue!6}$\mathbf{87.74}$ \popegain{8.24}
& \cellcolor{cvprblue!6}$\mathbf{85.80}$ \popegain{3.45}
\tabularnewline

\midrule

\multirow{5}{*}{\rotatebox[origin=c]{90}{\textbf{A-OKVQA}}}
& Regular
& $74.81$
& $68.81$
& $78.83$
& $74.89$
& $69.80$
& $78.24$
& $76.71$
& $70.90$
& $80.01$
\tabularnewline

& VCD~\cite{leng2024mitigating}
& $74.54$ \popeloss{0.27}
& $68.27$ \popeloss{0.54}
& $78.80$ \popeloss{0.03}
& $76.37$ \popegain{1.48}
& $71.57$ \popegain{1.77}
& $79.14$ \popegain{0.90}
& $76.90$ \popegain{0.19}
& $71.06$ \popegain{0.16}
& $80.24$ \popegain{0.23}
\tabularnewline

& AvisC~\cite{woo2024attentional}
& $77.47$ \popegain{2.66}
& $71.32$ \popegain{2.51}
& $80.92$ \popegain{2.09}
& $\underline{80.00}$ \popegain{5.11}
& $\underline{75.87}$ \popegain{6.07}
& $\underline{82.16}$ \popegain{3.92}
& $\underline{81.01}$ \popegain{4.30}
& $\underline{76.07}$ \popegain{5.17}
& $\underline{83.13}$ \popegain{3.12}
\tabularnewline

& ONLY~\cite{wan2025only}
& $\underline{77.94}$ \popegain{3.13}
& $\underline{71.79}$ \popegain{2.98}
& $\underline{81.38}$ \popegain{2.55}
& $78.58$ \popegain{3.69}
& $72.76$ \popegain{2.96}
& $81.66$ \popegain{3.42}
& $76.29$ \popeloss{0.42}
& $69.76$ \popeloss{1.14}
& $80.36$ \popegain{0.35}
\tabularnewline

& \cellcolor{cvprblue!6}Ours
& \cellcolor{cvprblue!6}$\mathbf{79.40}$ \popegain{4.59}
& \cellcolor{cvprblue!6}$\mathbf{73.98}$ \popegain{5.17}
& \cellcolor{cvprblue!6}$\mathbf{82.17}$ \popegain{3.34}
& \cellcolor{cvprblue!6}$\mathbf{82.11}$ \popegain{7.22}
& \cellcolor{cvprblue!6}$\mathbf{80.14}$ \popegain{10.34}
& \cellcolor{cvprblue!6}$\mathbf{83.14}$ \popegain{4.90}
& \cellcolor{cvprblue!6}$\mathbf{81.40}$ \popegain{4.69}
& \cellcolor{cvprblue!6}$\mathbf{77.07}$ \popegain{6.17}
& \cellcolor{cvprblue!6}$\mathbf{83.30}$ \popegain{3.29}
\tabularnewline

\midrule

\multirow{5}{*}{\rotatebox[origin=c]{90}{\textbf{GQA}}}
& Regular
& $73.96$
& $67.82$
& $78.36$
& $73.59$
& $68.88$
& $76.92$
& $75.87$
& $69.68$
& $79.56$
\tabularnewline

& VCD~\cite{leng2024mitigating}
& $73.65$ \popeloss{0.31}
& $67.11$ \popeloss{0.71}
& $78.43$ \popegain{0.07}
& $74.46$ \popegain{0.87}
& $70.14$ \popegain{1.26}
& $77.35$ \popegain{0.43}
& $76.13$ \popegain{0.26}
& $69.79$ \popegain{0.11}
& $79.90$ \popegain{0.34}
\tabularnewline

& AvisC~\cite{woo2024attentional}
& $76.49$ \popegain{2.53}
& $69.82$ \popegain{2.00}
& $80.55$ \popegain{2.19}
& $\underline{78.48}$ \popegain{4.89}
& $\underline{74.72}$ \popegain{5.84}
& $\underline{80.66}$ \popegain{3.74}
& $\underline{80.49}$ \popegain{4.62}
& $\underline{74.75}$ \popegain{5.07}
& $\underline{82.95}$ \popegain{3.39}
\tabularnewline

& ONLY~\cite{wan2025only}
& $\underline{76.62}$ \popegain{2.66}
& $\underline{70.02}$ \popegain{2.20}
& $\underline{80.72}$ \popegain{2.36}
& $77.02$ \popegain{3.43}
& $71.35$ \popegain{2.47}
& $80.26$ \popegain{3.34}
& $75.04$ \popeloss{0.83}
& $68.15$ \popeloss{1.53}
& $79.69$ \popegain{0.13}
\tabularnewline

& \cellcolor{cvprblue!6}Ours
& \cellcolor{cvprblue!6}$\mathbf{78.31}$ \popegain{4.35}
& \cellcolor{cvprblue!6}$\mathbf{72.30}$ \popegain{4.48}
& \cellcolor{cvprblue!6}$\mathbf{81.65}$ \popegain{3.29}
& \cellcolor{cvprblue!6}$\mathbf{80.36}$ \popegain{6.77}
& \cellcolor{cvprblue!6}$\mathbf{79.45}$ \popegain{10.57}
& \cellcolor{cvprblue!6}$\mathbf{81.00}$ \popegain{4.08}
& \cellcolor{cvprblue!6}$\mathbf{80.78}$ \popegain{4.91}
& \cellcolor{cvprblue!6}$\mathbf{75.62}$ \popegain{5.94}
& \cellcolor{cvprblue!6}$\mathbf{83.06}$ \popegain{3.50}
\tabularnewline

\bottomrule
\end{tabularx}
\endgroup

\vspace{-5pt}
\end{table*}

\noindent\textbf{Adaptive prediction fusion.}
At step $t$, the branches yield logits $\mathbf{z}_t^m$ and $\mathbf{z}_t^a$ and distributions $\mathbf{p}_t^m$ and $\mathbf{p}_t^a$. Following ONLY~\cite{wan2025only}, we compute their total variation distance as $D_t=\frac{1}{2}\sum_{v\in\mathcal{V}}|p_t^m(v)-p_t^a(v)|\in[0,1]$ and fuse the logits as:
\begin{equation}
    \mathbf{z}_t=
    \begin{cases}
        \mathbf{z}_t^m+\lambda_{+}\mathbf{z}_t^a,
        & D_t<\gamma,\\
        (1+\lambda_{-})\mathbf{z}_t^m-\lambda_{-}\mathbf{z}_t^a,
        & D_t\geq\gamma,
    \end{cases}
    \label{eq:adaptive-fusion}
\end{equation}
where $\gamma$ is the switching threshold, and $\lambda_{+}$ and $\lambda_{-}$ control collaborative and contrastive fusion, respectively. When the branches agree, additive fusion retains their shared and complementary predictions. A large discrepancy instead activates contrastive subtraction, favoring tokens supported by the intact evidence. A plausibility constraint~\cite{li2023contrastive,leng2024mitigating} filters candidates with threshold $\beta$. The chosen token is then appended to both branches, keeping their textual context aligned for the next step.

\section{Experiments}
\label{sec:experiments}

\begin{table*}[t]
    \centering
    
    \caption{CHAIR~\citep{rohrbach2018object} results at maximum generation lengths of 64 and 128 tokens. CHAIR$_S$ and CHAIR$_I$ denote sentence- and instance-level hallucination; Recall and Length measure object coverage and average caption length.}
    \label{tab:CHAIR}
    
    \begingroup
    \scriptsize
    \setlength{\tabcolsep}{2.2pt}
    \renewcommand{\arraystretch}{0.92}
    
    \begin{tabularx}{\linewidth}{
        >{\centering\arraybackslash}p{0.70cm}
        >{\raggedright\arraybackslash}p{1.20cm}
        *{4}{>{\centering\arraybackslash}X}
        @{\hspace{2.0pt}}
        !{\color{black!25}\vrule width 0.45pt}
        @{\hspace{2.0pt}}
        *{4}{>{\centering\arraybackslash}X}
        @{\hspace{2.0pt}}
        !{\color{black!25}\vrule width 0.45pt}
        @{\hspace{2.0pt}}
        *{4}{>{\centering\arraybackslash}X}
    }
    
    \toprule
    
    &
    \multirow{2}{*}{\textbf{Method}}
    & \multicolumn{4}{c}{\textbf{LLaVA-1.5-7B}}
    & \multicolumn{4}{c}{\textbf{InstructBLIP-7B}}
    & \multicolumn{4}{c}{\textbf{ShareGPT4V-7B}}
    \\
    
    \cmidrule(lr){3-6}
    \cmidrule(lr){7-10}
    \cmidrule(lr){11-14}
    
    &
    & \mbox{CHAIR$_S$ $\downarrow$}
    & \mbox{CHAIR$_I$ $\downarrow$}
    & \mbox{Recall $\uparrow$}
    & \mbox{Length $\uparrow$}
    
    & \mbox{CHAIR$_S$ $\downarrow$}
    & \mbox{CHAIR$_I$ $\downarrow$}
    & \mbox{Recall $\uparrow$}
    & \mbox{Length $\uparrow$}
    
    & \mbox{CHAIR$_S$ $\downarrow$}
    & \mbox{CHAIR$_I$ $\downarrow$}
    & \mbox{Recall $\uparrow$}
    & \mbox{Length $\uparrow$}
    \\
    
    \midrule
    
    
    \multirow{5}{*}{\rotatebox[origin=c]{90}{64 Tokens}}
    & Regular
    & $25.2$ & $8.8$ & $55.9$ & $53.6$
    & $30.6$ & $11.6$ & $55.9$ & $\underline{54.1}$
    & $16.0$ & $7.4$ & $42.3$ & $\underline{52.0}$
    \\
    
    & VCD~\cite{leng2024mitigating}
    & $23.4$ & $7.6$ & $\underline{61.1}$ & $\underline{54.1}$
    & $30.0$ & $10.1$ & $\underline{59.7}$ & $\mathbf{54.4}$
    & $17.0$ & $8.1$ & $\underline{43.1}$ & $\underline{52.0}$
    \\
    
    & AvisC~\cite{woo2024attentional}
    & $\underline{22.0}$ & $7.7$ & $58.6$ & $51.8$
    & $35.4$ & $11.2$ & $\mathbf{60.5}$ & $53.6$
    & $16.4$ & $8.8$ & $40.0$ & $50.0$
    \\
    
    & ONLY~\cite{wan2025only}
    & $22.4$ & $\underline{7.5}$ & $60.4$ & $53.2$
    & $\underline{26.8}$ & $\underline{9.4}$ & $58.8$ & $51.6$
    & $\underline{14.6}$ & $\underline{6.7}$ & $\mathbf{43.3}$ & $51.4$
    \\
    
    &
    \cellcolor{cvprblue!6}Ours
    & \cellcolor{cvprblue!6}$\mathbf{20.6}$
    & \cellcolor{cvprblue!6}$\mathbf{6.5}$
    & \cellcolor{cvprblue!6}$\mathbf{62.4}$
    & \cellcolor{cvprblue!6}$\mathbf{54.6}$
    & \cellcolor{cvprblue!6}$\mathbf{23.0}$
    & \cellcolor{cvprblue!6}$\mathbf{8.6}$
    & \cellcolor{cvprblue!6}$58.6$
    & \cellcolor{cvprblue!6}$53.6$
    & \cellcolor{cvprblue!6}$\mathbf{12.2}$
    & \cellcolor{cvprblue!6}$\mathbf{5.7}$
    & \cellcolor{cvprblue!6}$41.4$
    & \cellcolor{cvprblue!6}$\mathbf{52.7}$
    \\
    
    \midrule
    
    
    \multirow{5}{*}{\rotatebox[origin=c]{90}{128 Tokens}}
    & Regular
    & $\underline{53.0}$ & $15.6$ & $71.1$ & $97.6$
    & $\underline{56.6}$ & $\underline{16.4}$ & $68.0$ & $\underline{101.1}$
    & $\underline{29.0}$ & $9.8$ & $54.2$ & $103.0$
    \\
    
    & VCD~\cite{leng2024mitigating}
    & $58.8$ & $16.0$ & $\underline{75.9}$ & $97.7$
    & $56.8$ & $16.8$ & $\mathbf{72.3}$ & $100.4$
    & $31.2$ & $10.3$ & $\mathbf{57.3}$ & $\underline{104.0}$
    \\
    
    & AvisC~\cite{woo2024attentional}
    & $58.4$ & $16.7$ & $75.2$ & $\underline{99.0}$
    & $71.6$ & $20.1$ & $70.2$ & $97.5$
    & $31.0$ & $10.7$ & $55.7$ & $99.9$
    \\
    
    & ONLY~\cite{wan2025only}
    & $54.6$ & $\underline{15.1}$ & $74.9$ & $96.0$
    & $57.4$ & $17.5$ & $\underline{70.5}$ & $99.2$
    & $29.2$ & $\underline{9.0}$ & $\underline{56.8}$ & $102.7$
    \\
    
    &
    \cellcolor{cvprblue!6}Ours
    & \cellcolor{cvprblue!6}$\mathbf{48.4}$
    & \cellcolor{cvprblue!6}$\mathbf{14.3}$
    & \cellcolor{cvprblue!6}$\mathbf{78.6}$
    & \cellcolor{cvprblue!6}$\mathbf{100.8}$
    & \cellcolor{cvprblue!6}$\mathbf{46.0}$
    & \cellcolor{cvprblue!6}$\mathbf{14.0}$
    & \cellcolor{cvprblue!6}$69.0$
    & \cellcolor{cvprblue!6}$\mathbf{101.6}$
    & \cellcolor{cvprblue!6}$\mathbf{24.2}$
    & \cellcolor{cvprblue!6}$\mathbf{7.6}$
    & \cellcolor{cvprblue!6}$54.6$
    & \cellcolor{cvprblue!6}$\mathbf{105.0}$
    \\
    
    \bottomrule
    
    \end{tabularx}
    
    \endgroup
    
    \vspace{-5pt}
    
    \end{table*}

\definecolor{MMEGain}{rgb}{0.25,0.49,0.41}
\definecolor{MMELoss}{rgb}{0.71,0.37,0.37}
\newcommand{\mmegain}[1]{%
    \textcolor{MMEGain}{\tiny\,(+#1)}%
}
\newcommand{\mmeloss}[1]{%
    \textcolor{MMELoss}{\tiny\,(-#1)}%
}

\begin{table}[t]
\centering

\caption{MME-Hallucination~\citep{fu2025mme} results across three backbones. Scores are reported for object- and attribute-level subsets; colored values show absolute changes from Regular decoding.}
\label{tab:MME}

\begingroup
\scriptsize
\setlength{\tabcolsep}{2.0pt}
\renewcommand{\arraystretch}{0.94}

\begin{tabularx}{\linewidth}{
    @{}
    >{\raggedright\arraybackslash}p{0.82cm}
    *{4}{>{\centering\arraybackslash}X}
    >{\centering\arraybackslash}p{1.62cm}
    @{}
}

\toprule

\multirow{2}{*}{\textbf{Method}}
& \multicolumn{2}{c}{\textbf{Object-level}}
& \multicolumn{2}{c}{\textbf{Attribute-level}}
& \multirow{2}{*}{\makecell{\textbf{MME Score}\\$\uparrow$}}
\\

\cmidrule(lr){2-3}
\cmidrule(lr){4-5}

&
\mbox{Existence $\uparrow$}
&
\mbox{Count $\uparrow$}
&
\mbox{Position $\uparrow$}
&
\mbox{Color $\uparrow$}
&
\\

\midrule


\rowcolor{black!4}
\multicolumn{6}{c}{\textbf{LLaVA-1.5-7B}}
\\
\midrule

Regular
& $\underline{185.00}$
& $116.67$
& $\underline{128.33}$
& $160.00$
& $590.00$
\\

VCD~\cite{leng2024mitigating}
& $\underline{185.00}$
& $118.33$
& $126.67$
& $\underline{163.33}$
& \mbox{$593.33$\mmegain{3.33}}
\\

AvisC~\cite{woo2024attentional}
& $\mathbf{190.00}$
& $136.67$
& $106.67$
& $155.00$
& \mbox{$588.33$\mmeloss{1.67}}
\\

ONLY~\cite{wan2025only}
& $\mathbf{190.00}$
& $\underline{141.67}$
& $\underline{128.33}$
& $\mathbf{165.00}$
& \mbox{$\underline{625.00}$\mmegain{35.00}}
\\

\rowcolor{cvprblue!6}
Ours
& $\mathbf{190.00}$
& $\mathbf{158.33}$
& $\mathbf{141.67}$
& $160.00$
& \mbox{$\mathbf{650.00}$\mmegain{60.00}}
\\

\midrule


\rowcolor{black!4}
\multicolumn{6}{c}{\textbf{InstructBLIP-7B}}
\\
\midrule

Regular
& $170.00$
& $\mathbf{76.67}$
& $45.00$
& $118.33$
& $410.00$
\\

VCD~\cite{leng2024mitigating}
& $\mathbf{195.00}$
& $65.00$
& $\underline{65.00}$
& $118.33$
& \mbox{$\underline{443.33}$\mmegain{33.33}}
\\

AvisC~\cite{woo2024attentional}
& $\underline{185.00}$
& $60.00$
& $48.33$
& $\underline{126.67}$
& \mbox{$420.00$\mmegain{10.00}}
\\

ONLY~\cite{wan2025only}
& $\mathbf{195.00}$
& $55.00$
& $\underline{65.00}$
& $108.33$
& \mbox{$423.33$\mmegain{13.33}}
\\

\rowcolor{cvprblue!6}
Ours
& $180.00$
& $\underline{70.00}$
& $\mathbf{68.33}$
& $\mathbf{135.00}$
& \mbox{$\mathbf{453.33}$\mmegain{43.33}}
\\

\midrule


\rowcolor{black!4}
\multicolumn{6}{c}{\textbf{ShareGPT4V-7B}}
\\
\midrule

Regular
& $\underline{185.00}$
& $101.67$
& $106.67$
& $178.33$
& $571.67$
\\

VCD~\cite{leng2024mitigating}
& $\underline{185.00}$
& $131.67$
& $116.67$
& $\mathbf{185.00}$
& \mbox{$618.34$\mmegain{46.67}}
\\

AvisC~\cite{woo2024attentional}
& $180.00$
& $\underline{148.33}$
& $\underline{121.67}$
& $175.00$
& \mbox{$625.00$\mmegain{53.33}}
\\

ONLY~\cite{wan2025only}
& $\mathbf{195.00}$
& $140.00$
& $116.67$
& $\underline{180.00}$
& \mbox{$\underline{631.67}$\mmegain{60.00}}
\\

\rowcolor{cvprblue!6}
Ours
& $\mathbf{195.00}$
& $\mathbf{150.00}$
& $\mathbf{141.67}$
& $\mathbf{185.00}$
& \mbox{$\mathbf{671.67}$\mmegain{100.00}}
\\

\bottomrule

\end{tabularx}
\endgroup

\vspace{-5pt}
\end{table}

\subsection{Datasets and Evaluation Metrics}
\label{sec:datasets-metrics}
We evaluate VEGD on three complementary hallucination benchmarks. CHAIR~\cite{rohrbach2018object} measures hallucinated objects in captions of MS-COCO images~\cite{lin2014microsoft}. We report sentence-level CHAIR$_S$ and instance-level CHAIR$_I$, where lower values are better, together with object Recall and average caption Length to account for description coverage. CHAIR uses 500 images with maximum generation lengths of 64 and 128 tokens. POPE~\cite{li2023evaluating} contains balanced yes/no questions derived from MS-COCO~\cite{lin2014microsoft}, A-OKVQA~\cite{schwenk2022aokvqa}, and GQA~\cite{hudson2019gqa}. For each source, we use the full random, popular, and adversarial splits of 3,000 questions and report accuracy, precision, and F1. MME-Hallucination~\cite{fu2025mme} contains existence, count, position, and color questions. We report the official score for each subset and their sum, with higher values indicating better performance.

\subsection{Implementation Details}
\label{sec:implementation-details}
We evaluate eight LVLMs spanning established and recent architectures. The established backbones are LLaVA-1.5-7B~\cite{liu2023visual}, InstructBLIP-7B~\cite{dai2023instructblip}, and ShareGPT4V-7B~\cite{chen2024sharegpt4v}. The recent backbones are DeepSeek-VL2-Small~\cite{wu2024deepseekvl2}, Janus-Pro-7B~\cite{chen2025januspro}, Kimi-VL-A3B-Instruct~\cite{kimi2025kimivl}, InternVL3-9B~\cite{zhu2025internvl3}, and Qwen2.5-VL-7B~\cite{bai2025qwen25vl}. For a fair comparison, we reimplement all compared methods in the same evaluation framework and use identical generation settings for each model and benchmark. Visual evidence-guided auxiliary branch construction uses $\mathcal{L}_{\mathrm{loc}}=\{10,\ldots,20\}$ and selects $K=6$ localization heads, whose maps are smoothed with a $7\times7$ Gaussian kernel with $\sigma=1$. For Spatial Attention Intervention, we set $\mathcal{L}_{\mathrm{anc}}$ to the final 7 decoder layers, retain $K_a=16$ heads per layer, set $q^{-}$ and $q^{+}$ to the 0.1 and 0.9 quantiles of the aligned evidence scores, and use $\rho=3$ and $\alpha=3$ for both branches. Adaptive prediction fusion uses $\lambda_{+}=3$, $\lambda_{-}=1$, $\beta=0.4$, and $\gamma=0.5$. We use multinomial sampling over the full next-token distribution and set the temperature to 1.0 throughout. All experiments run on NVIDIA A100 40GB GPUs.

\subsection{Main Results}
\label{sec:main-results}

\noindent\textbf{Results on POPE.}
As shown in Tab.~\ref{tab:POPE}, VEGD achieves the best averaged accuracy across all nine dataset and backbone combinations while improving F1 over regular decoding in every case. It also gives the highest precision in all nine settings, indicating fewer false-positive object claims. Its averaged-accuracy margin over the strongest competing method reaches 2.11, with consistent gains across MS-COCO, A-OKVQA, and GQA.

\noindent\textbf{Results on CHAIR.}
As shown in Tab.~\ref{tab:CHAIR}, VEGD achieves the lowest CHAIR$_S$ and CHAIR$_I$ across backbones and generation lengths. The reductions remain clear at 128 tokens, where CHAIR$_S$ decreases from 53.0 to 48.4 on LLaVA, from 56.6 to 46.0 on InstructBLIP, and from 29.0 to 24.2 on ShareGPT4V. Across both generation budgets, recall improves on most backbones and captions remain comparable or longer, showing that lower hallucination does not come from reduced description coverage.

\noindent\textbf{Results on MME.}
As shown in Tab.~\ref{tab:MME}, VEGD achieves the highest overall MME score on all three backbones, reaching 650.00 on LLaVA, 453.33 on InstructBLIP, and 671.67 on ShareGPT4V. Compared with regular decoding, these totals increase by 60.00, 43.33, and 100.00, respectively. Most of the aggregate improvement across these models comes from count and position, showing that the benefit extends beyond object-existence verification.

\noindent\textbf{Results on Modern Backbones.}
As shown in Tab.~\ref{tab:chair-modern-backbones-temp1}, VEGD lowers CHAIR$_S$ and CHAIR$_I$ and raises recall across all 10 model and length settings on five recent LVLMs. Across these settings, the mean reductions in CHAIR$_S$ and CHAIR$_I$ are 3.56 and 1.61, with a 2.32 increase in recall. On DeepSeek-VL2-Small at 128 tokens, VEGD reduces CHAIR$_S$ from 33.6 to 28.4 and increases recall from 61.40 to 64.78. Caption length also increases in most settings, demonstrating transfer beyond the established backbones without sacrificing description coverage.

\begin{table}[t]
    \centering
    
    \caption{CHAIR comparison of Regular decoding and VEGD across five recent LVLMs. MT denotes maximum output tokens; C$_S$ and C$_I$ denote sentence- and instance-level hallucination; R and Len denote object recall and average caption length.}
    \label{tab:chair-modern-backbones-temp1}
    
    \begingroup
    \scriptsize
    \setlength{\tabcolsep}{2.0pt}
    \renewcommand{\arraystretch}{0.94}
    
    \begin{tabularx}{\linewidth}{
        @{}
        >{\raggedright\arraybackslash}p{2.28cm}
        @{}
        >{\hspace{3pt}\raggedright\arraybackslash}p{1.02cm}
        >{\centering\arraybackslash}p{0.46cm}
        *{4}{>{\centering\arraybackslash}X}
        @{}
    }
    
    \toprule
    \textbf{Backbone}
    & \textbf{Method}
    & \textbf{MT}
    & \mbox{C$_S$ $\downarrow$}
    & \mbox{C$_I$ $\downarrow$}
    & \mbox{R $\uparrow$}
    & \mbox{Len $\uparrow$}
    \\
    \midrule
    
    & Regular & $64$
    & $20.40$ & $7.92$ & $50.77$ & $59.38$
    \\
    
    \rowcolor{cvprblue!6}
    \cellcolor{white}
    & Ours & $64$
    & $\mathbf{15.60}$ & $\mathbf{5.79}$ & $\mathbf{52.18}$ & $\mathbf{60.58}$
    \\
    
    \cmidrule(lr){2-7}
    
    & Regular & $128$
    & $33.60$ & $9.70$ & $61.40$ & $104.27$
    \\
    
    \rowcolor{cvprblue!6}
    \cellcolor{white}
    \multirow{-4}{*}{\mbox{DeepSeek-VL2-Small}}
    & Ours & $128$
    & $\mathbf{28.40}$ & $\mathbf{8.27}$ & $\mathbf{64.78}$ & $\mathbf{106.28}$
    \\
    
    \midrule
    
    & Regular & $64$
    & $17.00$ & $6.54$ & $48.56$ & $60.67$
    \\
    
    \rowcolor{cvprblue!6}
    \cellcolor{white}
    & Ours & $64$
    & $\mathbf{12.20}$ & $\mathbf{4.53}$ & $\mathbf{50.40}$ & $\mathbf{60.78}$
    \\
    
    \cmidrule(lr){2-7}
    
    & Regular & $128$
    & $30.40$ & $9.05$ & $60.30$ & $\mathbf{109.16}$
    \\
    
    \rowcolor{cvprblue!6}
    \cellcolor{white}
    \multirow{-4}{*}{\mbox{Janus-Pro-7B}}
    & Ours & $128$
    & $\mathbf{27.60}$ & $\mathbf{8.45}$ & $\mathbf{62.82}$ & $103.87$
    \\
    
    \midrule
    
    & Regular & $64$
    & $19.00$ & $7.33$ & $54.70$ & $56.85$
    \\
    
    \rowcolor{cvprblue!6}
    \cellcolor{white}
    & Ours & $64$
    & $\mathbf{15.40}$ & $\mathbf{4.32}$ & $\mathbf{59.50}$ & $\mathbf{60.26}$
    \\
    
    \cmidrule(lr){2-7}
    
    & Regular & $128$
    & $26.00$ & $8.68$ & $58.33$ & $85.34$
    \\
    
    \rowcolor{cvprblue!6}
    \cellcolor{white}
    \multirow{-4}{*}{\mbox{Kimi-VL-A3B-Instruct}}
    & Ours & $128$
    & $\mathbf{23.20}$ & $\mathbf{5.65}$ & $\mathbf{62.51}$ & $\mathbf{97.23}$
    \\
    
    \midrule
    
    & Regular & $64$
    & $16.20$ & $6.70$ & $48.37$ & $53.41$
    \\
    
    \rowcolor{cvprblue!6}
    \cellcolor{white}
    & Ours & $64$
    & $\mathbf{14.60}$ & $\mathbf{5.73}$ & $\mathbf{49.85}$ & $\mathbf{54.10}$
    \\
    
    \cmidrule(lr){2-7}
    
    & Regular & $128$
    & $31.80$ & $9.55$ & $64.17$ & $103.30$
    \\
    
    \rowcolor{cvprblue!6}
    \cellcolor{white}
    \multirow{-4}{*}{\mbox{InternVL3-9B}}
    & Ours & $128$
    & $\mathbf{31.20}$ & $\mathbf{8.87}$ & $\mathbf{65.09}$ & $\mathbf{104.41}$
    \\
    
    \midrule
    
    & Regular & $64$
    & $17.20$ & $6.78$ & $42.04$ & $60.44$
    \\
    
    \rowcolor{cvprblue!6}
    \cellcolor{white}
    & Ours & $64$
    & $\mathbf{12.80}$ & $\mathbf{5.40}$ & $\mathbf{44.07}$ & $\mathbf{60.93}$
    \\
    
    \cmidrule(lr){2-7}
    
    & Regular & $128$
    & $31.20$ & $8.56$ & $56.79$ & $120.12$
    \\
    
    \rowcolor{cvprblue!6}
    \cellcolor{white}
    \multirow{-4}{*}{\mbox{Qwen2.5-VL-7B}}
    & Ours & $128$
    & $\mathbf{26.20}$ & $\mathbf{7.68}$ & $\mathbf{57.41}$ & $\mathbf{121.75}$
    \\
    
    \bottomrule
    
    \end{tabularx}
    \endgroup
    
    \vspace{-5pt}
    \end{table}

\subsection{Ablation Studies}
\label{sec:ablation-studies}

\begin{table}[t]
    \centering
    \caption{Component ablation on LLaVA-1.5-7B. ABC denotes auxiliary construction, while $\mathrm{SAI}_{m}$ and $\mathrm{SAI}_{a}$ denote Spatial Attention Intervention in the main and auxiliary branches. POPE refers to the adversarial POPE-COCO split.}
    \label{tab:component-ablation}
    \begingroup
    \scriptsize
    \setlength{\tabcolsep}{1.2pt}
    \renewcommand{\arraystretch}{0.98}
    \begin{tabularx}{\linewidth}{@{}*{8}{>{\centering\arraybackslash}X}@{}}
        \toprule
        \multicolumn{3}{c}{\textbf{Components}}
        & \multicolumn{2}{c}{\textbf{POPE}}
        & \multicolumn{2}{c}{\textbf{CHAIR-64 Tokens}}
        & \multirow{2}{*}{\makecell{\textbf{MME}\\\textbf{Score} $\uparrow$}} \\
        \cmidrule(lr){1-3}\cmidrule(lr){4-5}\cmidrule(lr){6-7}
        \textbf{ABC} & $\mathbf{SAI}_{m}$ & $\mathbf{SAI}_{a}$
        & \textbf{Acc.}$\uparrow$ & \textbf{F1}$\uparrow$
        & $\mathbf{C}_S\downarrow$ & $\mathbf{C}_I\downarrow$ & \\
        \midrule
        & & & $75.93$ & $78.33$ & $25.2$ & $8.8$ & $590.00$ \\
        $\checkmark$ & & & $80.37$ & $81.73$ & $21.4$ & $6.8$ & $\underline{621.67}$ \\
        $\checkmark$ & $\checkmark$ & & $\underline{81.07}$ & $\underline{82.01}$ & $\underline{20.6}$ & $6.7$ & $596.67$ \\
        $\checkmark$ & & $\checkmark$ & $80.73$ & $82.00$ & $\mathbf{19.8}$ & $\mathbf{6.1}$ & $606.67$ \\
        \rowcolor{cvprblue!6}
        $\checkmark$ & $\checkmark$ & $\checkmark$
        & $\mathbf{81.43}$ & $\mathbf{82.24}$ & $\underline{20.6}$ & $\underline{6.5}$ & $\mathbf{650.00}$ \\
        \bottomrule
    \end{tabularx}
    \endgroup
\end{table}

\noindent\textbf{Component analysis.}
Tab.~\ref{tab:component-ablation} evaluates visual evidence-guided auxiliary-image construction and Spatial Attention Intervention in the main and auxiliary branches, showing that auxiliary-image construction alone improves all three benchmarks over regular decoding. The one-sided variants favor different tasks, with main-branch intervention benefiting POPE and auxiliary-branch intervention performing best on CHAIR, but both reduce MME relative to auxiliary-image construction alone. Applying it to both branches yields the best POPE and MME results while keeping CHAIR below regular decoding and auxiliary-image construction alone. This balance supports guiding access to intact evidence in the main branch and its weakened counterpart in the auxiliary branch.

\begin{figure}[t]
    \centering
    \includegraphics[width=\linewidth]{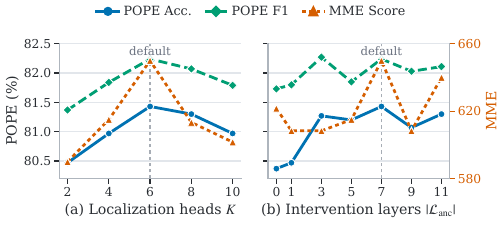}
    \caption{\textbf{Sensitivity to head count and intervention depth.} (a) varies $K$ with intervention in the final 7 layers. (b) varies $|\mathcal{L}_{\mathrm{anc}}|$ with $K=6$. Results are on adversarial POPE-COCO and MME-Hallucination. Vertical lines mark the defaults.}
    \label{fig:hyperparameter-sensitivity}
\end{figure}

\noindent\textbf{Hyperparameter sensitivity.}
Fig.~\ref{fig:hyperparameter-sensitivity} studies the localization-head count $K$ and intervention depth. Performance peaks at $K=6$. Adding more heads leaves POPE nearly unchanged but lowers MME, suggesting that extra heads add noisy localization cues. POPE also varies little with depth, whereas MME is more sensitive. The final 7 layers give the best POPE accuracy and MME score, with F1 only 0.03 below its peak. We therefore use $K=6$ and $|\mathcal{L}_{\mathrm{anc}}|=7$ by default.
Additional ablations are provided in the supplementary material.

\subsection{Analysis}
\label{sec:analysis}

\begin{figure}[t]
    \centering
    \includegraphics[width=\linewidth]{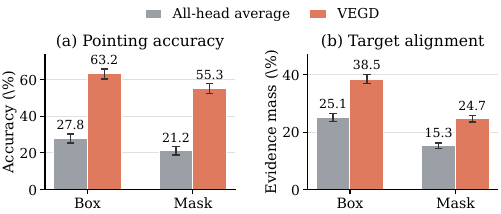}
    \caption{\textbf{Visual evidence localization.} Results on positive adversarial POPE-COCO queries with LLaVA-1.5-7B. COCO boxes and instance masks are used only for evaluation. Error bars denote 95\% bootstrap confidence intervals.}
    \label{fig:visual-evidence-localization}
\end{figure}

\begin{figure}[t]
    \centering
    \includegraphics[width=\linewidth]{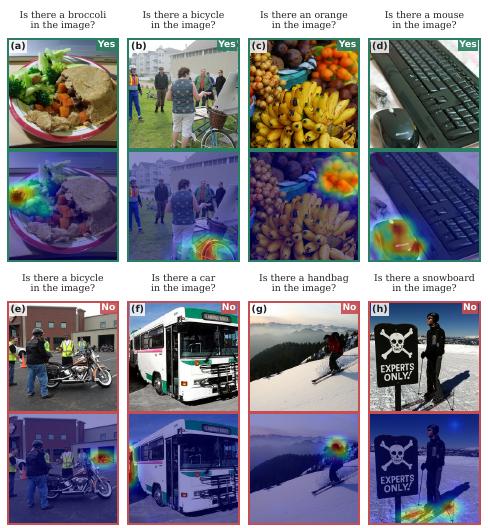}
    \caption{\textbf{Visual evidence for positive and negative queries.} Present targets are localized directly, while absent targets activate related candidates. Green/red frames denote Yes/No. Results are from LLaVA-1.5-7B on adversarial POPE-COCO.}
    \label{fig:visual-evidence-cases}
\end{figure}

\begin{figure}[t]
    \centering
    \includegraphics[width=\linewidth]{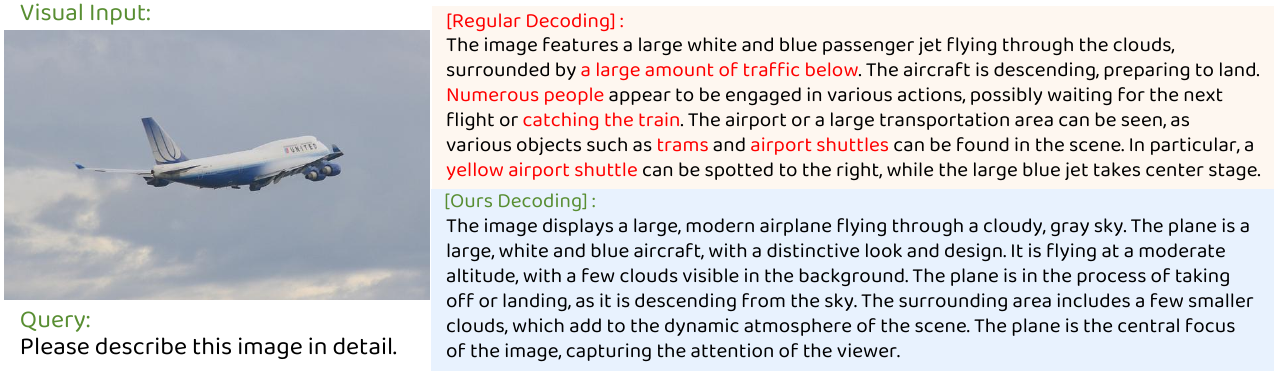}
    \caption{\textbf{Case Study.} Results are obtained with InstructBLIP-7B on CHAIR. Red text marks hallucinated content.}
    \label{fig:qualitative-chair}
\end{figure}

\noindent\textbf{Visual evidence localization.}
Fig.~\ref{fig:visual-evidence-localization} compares VEGD with all-head averaging using external COCO boxes and instance masks. VEGD raises pointing accuracy from 27.8 to 63.2 with boxes and from 21.2 to 55.3 with masks. Similar gains in evidence mass show that selecting localization heads preserves more query-relevant spatial information than aggregating all heads.
Fig.~\ref{fig:visual-evidence-cases} shows distinct responses to positive and negative queries. For positive queries, the map aligns with the present object. VEGD strengthens this evidence in the main branch and weakens the same region in the auxiliary branch, preserving visually supported Yes predictions. When the target is absent, the map shifts to a related candidate, such as a motorcycle for bicycle or skis for snowboard. Guiding both branches to that region retains the details needed to reject the candidate and exposes Yes scores without matching support. VEGD thus preserves true positives while suppressing hallucinated ones.

\noindent\textbf{Case study.}
As shown in Fig. ~\ref{fig:qualitative-chair}, regular decoding hallucinates people, trains, and airport vehicles in an image containing only an airplane and sky. VEGD avoids these errors with a response of similar length, showing that the gain does not come from shorter descriptions. More examples are provided in the supplementary material.

\section{Conclusion}
\label{sec:conclusion}

This work revisits training-free contrastive decoding for LVLM hallucination mitigation by examining how both branches use visual evidence. We introduced Visual Evidence-Guided Decoding (VEGD), which identifies query-relevant evidence through selected localization heads, weakens it to construct the auxiliary image, and anchors both branches to its location throughout decoding. Across CHAIR, POPE, and MME-Hallucination with eight LVLMs, VEGD reduces hallucinations and improves fine-grained perception while preserving description coverage. These results show that query-relevant visual evidence should guide contrastive decoding throughout generation, not just auxiliary-branch construction.

\section*{LLM Usage Statement}
OpenAI Codex with the GPT-5.6 Sol model assisted with language editing, manuscript organization, and stylistic refinement. GPT-5.6 Sol was also used to improve the clarity and internal consistency of technical descriptions in the paper and supplementary material. The authors developed the methodology, designed and conducted all experiments, analyzed the results, and independently determined the scientific interpretations. All AI-assisted manuscript text was reviewed and validated by the authors, with numerical statements checked against the underlying experimental outputs. The authors take full responsibility for the study design, reported results, interpretations, and final manuscript.

{
    \small
    \bibliographystyle{ieeenat_fullname}
    \bibliography{main}

@String(CVPR= {IEEE Conf. Comput. Vis. Pattern Recog.})

@String(ICCV= {Int. Conf. Comput. Vis.})

@String(ICASSP=	{ICASSP})

@String(CVPR  = {CVPR})

@String(ICCV  = {ICCV})

@inproceedings{liu2023visual,
  title={Improved baselines with visual instruction tuning},
  author={Liu, Haotian and Li, Chunyuan and Li, Yuheng and Lee, Yong Jae},
  booktitle={2024 IEEE/CVF Conference on Computer Vision and Pattern Recognition (CVPR)},
  pages={26286--26296},
  year={2024},
  organization={IEEE}
}

@article{dai2023instructblip,
  title={Instructblip: Towards general-purpose vision-language models with instruction tuning},
  author={Dai, Wenliang and Li, Junnan and Li, Dongxu and Tiong, Anthony and Zhao, Junqi and Wang, Weisheng and Li, Boyang and Fung, Pascale N and Hoi, Steven},
  journal={Advances in neural information processing systems},
  volume={36},
  pages={49250--49267},
  year={2023}
}

@inproceedings{li2023evaluating,
  title={Evaluating object hallucination in large vision-language models},
  author={Li, Yifan and Du, Yifan and Zhou, Kun and Wang, Jinpeng and Zhao, Wayne Xin and Wen, Ji-Rong},
  booktitle={Proceedings of the 2023 conference on empirical methods in natural language processing},
  pages={292--305},
  year={2023}
}

@inproceedings{leng2024mitigating,
  title={Mitigating object hallucinations in large vision-language models through visual contrastive decoding},
  author={Leng, Sicong and Zhang, Hang and Chen, Guanzheng and Li, Xin and Lu, Shijian and Miao, Chunyan and Bing, Lidong},
  booktitle={2024 IEEE/CVF Conference on Computer Vision and Pattern Recognition (CVPR)},
  pages={13872--13882},
  year={2024},
  organization={IEEE}
}

@article{huang2026chasd,
  title={CHASD: Language Increment-Calibrated Contrastive Decoding against Hallucination in LVLMs},
  author={Huang, Xiaoyi and Zhang, Kejia and Luo, Zhiming},
  journal={arXiv preprint arXiv:2605.23344},
  year={2026}
}

@inproceedings{liang2026envision,
  title={Envision, Attend, Then Respond: Counterfactual Hallucination Mitigation in Large Vision-Language Models},
  author={Liang, Yuxuan and Shi, Fan and Zhu, Rui and Li, Xu and Chen, Xiaolei and Liu, Zhe and Li, Bin and Xue, Xiangyang},
  booktitle={Proceedings of the IEEE/CVF Conference on Computer Vision and Pattern Recognition},
  pages={18261--18272},
  year={2026}
}

@article{woo2024ritual,
  title={Ritual: Random image transformations as a universal anti-hallucination lever in large vision language models},
  author={Woo, Sangmin and Jang, Jaehyuk and Kim, Donguk and Choi, Yubin and Kim, Changick},
  journal={arXiv preprint arXiv:2405.17821},
  year={2024}
}

@inproceedings{lee2026delve,
  title={Delve into visual contrastive decoding for hallucination mitigation of large vision-language models},
  author={Lee, Yi-Lun and Tsai, Yi-Hsuan and Chiu, Wei-Chen},
  booktitle={International Conference on Pattern Recognition},
  pages={175--189},
  year={2026},
  organization={Springer}
}

@inproceedings{favero2024multimodal,
  title={Multi-modal hallucination control by visual information grounding},
  author={Favero, Alessandro and Zancato, Luca and Trager, Matthew and Choudhary, Siddharth and Perera, Pramuditha and Achille, Alessandro and Swaminathan, Ashwin and Soatto, Stefano},
  booktitle={2024 IEEE/CVF Conference on Computer Vision and Pattern Recognition (CVPR)},
  pages={14303--14312},
  year={2024},
  organization={IEEE}
}

@inproceedings{liu2024paying,
  title={Paying more attention to image: A training-free method for alleviating hallucination in lvlms},
  author={Liu, Shi and Zheng, Kecheng and Chen, Wei},
  booktitle={European Conference on Computer Vision},
  pages={125--140},
  year={2024},
  organization={Springer}
}

@inproceedings{zhang2025selfcorrecting,
  title={Self-correcting decoding with generative feedback for mitigating hallucinations in large vision-language models},
  author={Zhang, Ce and Wan, Zifu and Kan, Zhehan and Ma, Martin Q and Stepputtis, Simon and Ramanan, Deva and Salakhutdinov, Russ and Morency, Louis-Philippe and Sycara, Katia and Xie, Yaqi},
  booktitle={International Conference on Learning Representations},
  volume={2025},
  pages={5494--5524},
  year={2025}
}

@inproceedings{woo2024attentional,
  title={Don’t miss the forest for the trees: Attentional vision calibration for large vision language models},
  author={Woo, Sangmin and Kim, Donguk and Jang, Jaehyuk and Choi, Yubin and Kim, Changick},
  booktitle={Findings of the Association for Computational Linguistics: ACL 2025},
  pages={1927--1951},
  year={2025}
}

@article{li2025hidden,
  title={The hidden life of tokens: Reducing hallucination of large vision-language models via visual information steering},
  author={Li, Zhuowei and Shi, Haizhou and Gao, Yunhe and Liu, Di and Wang, Zhenting and Chen, Yuxiao and Liu, Ting and Zhao, Long and Wang, Hao and Metaxas, Dimitris N},
  journal={arXiv preprint arXiv:2502.03628},
  year={2025}
}

@inproceedings{liu2025vti,
  title={Reducing hallucinations in large vision-language models via latent space steering},
  author={Liu, Sheng and Ye, Haotian and Zou, James Y},
  booktitle={International Conference on Learning Representations},
  volume={2025},
  pages={72402--72419},
  year={2025}
}

@inproceedings{an2025agla,
  title={Mitigating object hallucinations in large vision-language models with assembly of global and local attention},
  author={An, Wenbin and Tian, Feng and Leng, Sicong and Nie, Jiahao and Lin, Haonan and Wang, QianYing and Chen, Ping and Zhang, Xiaoqin and Lu, Shijian},
  booktitle={2025 IEEE/CVF Conference on Computer Vision and Pattern Recognition (CVPR)},
  pages={29915--29926},
  year={2025},
  organization={IEEE}
}

@inproceedings{wan2025only,
  title={Only: One-layer intervention sufficiently mitigates hallucinations in large vision-language models},
  author={Wan, Zifu and Zhang, Ce and Yong, Silong and Ma, Martin Q and Stepputtis, Simon and Morency, Louis-Philippe and Ramanan, Deva and Sycara, Katia and Xie, Yaqi},
  booktitle={2025 IEEE/CVF International Conference on Computer Vision (ICCV)},
  pages={3225--3234},
  year={2025},
  organization={IEEE}
}

@inproceedings{li2026cmac,
  title={Cross-modal attention calibration for lvlm hallucination mitigation},
  author={Li, Jiaming and Zhang, Jiacheng and Jie, Zequn and Ma, Lin and Li, Ming and Luo, Xiaonan and Li, Guanbin},
  booktitle={Proceedings of the IEEE/CVF Conference on Computer Vision and Pattern Recognition},
  pages={40186--40196},
  year={2026}
}

@inproceedings{huo2025self,
  title={Self-introspective decoding: Alleviating hallucinations for large vision-language models},
  author={Huo, Fushuo and Xu, Wenchao and Zhang, Zhong and Wang, Haozhao and Chen, Zhicheng and Zhao, Peilin},
  booktitle={International Conference on Learning Representations},
  volume={2025},
  pages={24272--24295},
  year={2025}
}

@inproceedings{rohrbach2018object,
  title={Object hallucination in image captioning},
  author={Rohrbach, Anna and Hendricks, Lisa Anne and Burns, Kaylee and Darrell, Trevor and Saenko, Kate},
  booktitle={Proceedings of the 2018 conference on empirical methods in natural language processing},
  pages={4035--4045},
  year={2018}
}

@inproceedings{fu2025mme,
 author = {Fu, Chaoyou and Chen, Peixian and Shen, Yunhang and Qin, Yulei and Zhang, Mengdan and Lin, Xu and Yang, Jinrui and Zheng, Xiawu and Li, Ke and Sun, Xing and Wu, Yunsheng and Ji, Rongrong and Shan, Caifeng and He, Ran},
 booktitle = {Advances in Neural Information Processing Systems},
 title = {MME: A Comprehensive Evaluation Benchmark for Multimodal Large Language Models},
 year = {2025}
}

@inproceedings{kang2025your,
  title={Your large vision-language model only needs a few attention heads for visual grounding},
  author={Kang, Seil and Kim, Jinyeong and Kim, Junhyeok and Hwang, Seong Jae},
  booktitle={2025 IEEE/CVF Conference on Computer Vision and Pattern Recognition (CVPR)},
  pages={9339--9350},
  year={2025},
  organization={IEEE}
}

@inproceedings{xie2026headaware,
  title={Head-Aware Visual Cropping: Enhancing Fine-Grained VQA with Attention-Guided Subimage},
  author={Xie, Junfei and Pan, Peng and Zhang, Xulong},
  booktitle={ICASSP 2026-2026 IEEE International Conference on Acoustics, Speech and Signal Processing (ICASSP)},
  pages={12402--12406},
  year={2026},
  organization={IEEE}
}

@inproceedings{wang2025sparsemm,
  title={Sparsemm: Head sparsity emerges from visual concept responses in mllms},
  author={Wang, Jiahui and Liu, Zuyan and Rao, Yongming and Lu, Jiwen},
  booktitle={2025 IEEE/CVF International Conference on Computer Vision (ICCV)},
  pages={23177--23187},
  year={2025},
  organization={IEEE}
}

@inproceedings{wu2024vstar,
  title={V*: Guided visual search as a core mechanism in multimodal llms},
  author={Wu, Penghao and Xie, Saining},
  booktitle={2024 IEEE/CVF Conference on Computer Vision and Pattern Recognition (CVPR)},
  pages={13084--13094},
  year={2024},
  organization={IEEE}
}

@inproceedings{zhang2025mllms,
  title={{MLLM}s Know Where to Look: Training-free Perception of Small Visual Details with Multimodal {LLM}s},
  author={Jiarui Zhang and Mahyar Khayatkhoei and Prateek Chhikara and Filip Ilievski},
  booktitle={The Thirteenth International Conference on Learning    Representations},
  year={2025},
  url={https://openreview.net/forum?id=DgaY5mDdmT}
}

@inproceedings{wu2025flmm,
  title={F-lmm: Grounding frozen large multimodal models},
  author={Wu, Size and Jin, Sheng and Zhang, Wenwei and Xu, Lumin and Liu, Wentao and Li, Wei and Loy, Chen Change},
  booktitle={2025 IEEE/CVF Conference on Computer Vision and Pattern Recognition (CVPR)},
  pages={24710--24721},
  year={2025},
  organization={IEEE}
}

@inproceedings{jiang2025devils,
  title={Devils in middle layers of large vision-language models: Interpreting, detecting and mitigating object hallucinations via attention lens},
  author={Jiang, Zhangqi and Chen, Junkai and Zhu, Beier and Luo, Tingjin and Shen, Yankun and Yang, Xu},
  booktitle={2025 IEEE/CVF Conference on Computer Vision and Pattern Recognition (CVPR)},
  pages={25004--25014},
  year={2025},
  organization={IEEE}
}

@article{alayrac2022flamingo,
  title={Flamingo: a visual language model for few-shot learning},
  author={Alayrac, Jean-Baptiste and Donahue, Jeff and Luc, Pauline and Miech, Antoine and Barr, Iain and Hasson, Yana and Lenc, Karel and Mensch, Arthur and Millican, Katherine and Reynolds, Malcolm and others},
  journal={Advances in neural information processing systems},
  volume={35},
  pages={23716--23736},
  year={2022}
}

@inproceedings{li2023blip2,
  title={Blip-2: Bootstrapping language-image pre-training with frozen image encoders and large language models},
  author={Li, Junnan and Li, Dongxu and Savarese, Silvio and Hoi, Steven},
  booktitle={International conference on machine learning},
  pages={19730--19742},
  year={2023},
  organization={PmLR}
}

@article{li2024llavaonevision,
  title={Llava-onevision: Easy visual task transfer},
  author={Li, Bo and Zhang, Yuanhan and Guo, Dong and Zhang, Renrui and Li, Feng and Zhang, Hao and Zhang, Kaichen and Zhang, Peiyuan and Li, Yanwei and Liu, Ziwei and others},
  journal={arXiv preprint arXiv:2408.03326},
  year={2024}
}

@article{bai2025qwen25vl,
  author        = {Bai, Shuai and Chen, Keqin and Liu, Xuejing and Wang, Jialin and Ge, Wenbin and Song, Sibo and others},
  title         = {{Qwen2.5-VL} Technical Report},
  journal        = {arXiv preprint arXiv:2502.13923},
  year          = {2025},
}

@article{zhu2025internvl3,
  title={Internvl3: Exploring advanced training and test-time recipes for open-source multimodal models},
  author={Zhu, Jinguo and Wang, Weiyun and Chen, Zhe and Liu, Zhaoyang and Ye, Shenglong and Gu, Lixin and Tian, Hao and Duan, Yuchen and Su, Weijie and Shao, Jie and others},
  journal={arXiv preprint arXiv:2504.10479},
  year={2025}
}

@inproceedings{zhu2024minigpt4,
  title={Minigpt-4: Enhancing vision-language understanding with advanced large language models},
  author = {Zhu, Deyao and Chen, Jun and Shen, Xiaoqian and Li, Xiang and Elhoseiny, Mohamed},
  booktitle={International Conference on Learning Representations},
  volume={2024},
  pages={18378--18394},
  year={2024}
}

@article{chen2023shikra,
  title={Shikra: Unleashing multimodal llm's referential dialogue magic},
  author={Chen, Keqin and Zhang, Zhao and Zeng, Weili and Zhang, Richong and Zhu, Feng and Zhao, Rui},
  journal={arXiv preprint arXiv:2306.15195},
  year={2023}
}

@inproceedings{li2023contrastive,
  title={Contrastive decoding: Open-ended text generation as optimization},
  author={Li, Xiang Lisa and Holtzman, Ari and Fried, Daniel and Liang, Percy and Eisner, Jason and Hashimoto, Tatsunori B and Zettlemoyer, Luke and Lewis, Mike},
  booktitle={Proceedings of the 61st annual meeting of the association for computational linguistics (volume 1: Long papers)},
  pages={12286--12312},
  year={2023}
}

@inproceedings{wang2024instruction,
  title={Mitigating hallucinations in large vision-language models with instruction contrastive decoding},
  author={Wang, Xintong and Pan, Jingheng and Ding, Liang and Biemann, Chris},
  booktitle={Findings of the association for computational linguistics: ACL 2024},
  pages={15840--15853},
  year={2024}
}

@InProceedings{jo2026attention,
    author    = {Jo, Yujin and Bae, Sangyoon and Kim, Taesup},
    title     = {Attention-Space Contrastive Guidance for Efficient Hallucination Mitigation in LVLMs},
    booktitle = {Proceedings of the IEEE/CVF Conference on Computer Vision and Pattern Recognition (CVPR) Findings},
    month     = {June},
    year      = {2026},
    pages     = {9706-9715}
}

@inproceedings{lin2014microsoft,
  title={Microsoft coco: Common objects in context},
  author={Lin, Tsung-Yi and Maire, Michael and Belongie, Serge and Hays, James and Perona, Pietro and Ramanan, Deva and Doll{\'a}r, Piotr and Zitnick, C Lawrence},
  booktitle={European conference on computer vision},
  pages={740--755},
  year={2014},
  organization={Springer}
}

@inproceedings{schwenk2022aokvqa,
  title={A-okvqa: A benchmark for visual question answering using world knowledge},
  author={Schwenk, Dustin and Khandelwal, Apoorv and Clark, Christopher and Marino, Kenneth and Mottaghi, Roozbeh},
  booktitle={European conference on computer vision},
  pages={146--162},
  year={2022},
  organization={Springer}
}

@inproceedings{hudson2019gqa,
  title={Gqa: A new dataset for real-world visual reasoning and compositional question answering},
  author={Hudson, Drew A and Manning, Christopher D},
  booktitle={2019 IEEE/CVF Conference on Computer Vision and Pattern Recognition (CVPR)},
  pages={6693--6702},
  year={2019},
  organization={IEEE}
}

@inproceedings{chen2024sharegpt4v,
  title={Sharegpt4v: Improving large multi-modal models with better captions},
  author={Chen, Lin and Li, Jinsong and Dong, Xiaoyi and Zhang, Pan and He, Conghui and Wang, Jiaqi and Zhao, Feng and Lin, Dahua},
  booktitle={European Conference on Computer Vision},
  pages={370--387},
  year={2024},
  organization={Springer}
}

@article{wu2024deepseekvl2,
  title={Deepseek-vl2: Mixture-of-experts vision-language models for advanced multimodal understanding},
  author={Wu, Zhiyu and Chen, Xiaokang and Pan, Zizheng and Liu, Xingchao and Liu, Wen and Dai, Damai and Gao, Huazuo and Ma, Yiyang and Wu, Chengyue and Wang, Bingxuan and others},
  journal={arXiv preprint arXiv:2412.10302},
  year={2024}
}

@article{chen2025januspro,
  title={Janus-pro: Unified multimodal understanding and generation with data and model scaling},
  author={Chen, Xiaokang and Wu, Zhiyu and Liu, Xingchao and Pan, Zizheng and Liu, Wen and Xie, Zhenda and Yu, Xingkai and Ruan, Chong},
  journal={arXiv preprint arXiv:2501.17811},
  year={2025}
}

@article{kimi2025kimivl,
  title={Kimi-vl technical report},
  author={Team, Kimi and Du, Angang and Yin, Bohong and Xing, Bowei and Qu, Bowen and Wang, Bowen and Chen, Cheng and Zhang, Chenlin and Du, Chenzhuang and Wei, Chu and others},
  journal={arXiv preprint arXiv:2504.07491},
  year={2025}
}

@inproceedings{lei2026see,
  title={See only when needed: Context-aware attention intervention for mitigating hallucinations in lvlms},
  author={Lei, Yuqing and Lyu, Wenbo and Du, Yingjun and Zhen, Xiantong and Snoek, Cees GM and Shao, Ling},
  booktitle={European Conference on Computer Vision},
  pages={154--177},
  year={2026},
  organization={Springer}
}
}


\end{document}